\documentclass[twocolumn,letterpaper]{IEEEAerospaceCLS}  %

\usepackage{adjustbox}
\usepackage{amsmath}
\usepackage{mathtools}
\usepackage{graphicx}
\usepackage{pgfgantt}
\usepackage{paralist}
\usepackage[bookmarks=false]{hyperref}
\usepackage{outlines}
\usepackage{float}
\usepackage[utf8]{inputenc}
\usepackage{url}
\usepackage{tabularx}
\usepackage{todonotes}
\usepackage[nolist]{acronym}
\usepackage{graphicx}    %
\newcommand{\ignore}[1]{}  %

\begin{document}
\title{Shapeshifter: A Multi-Agent, Multi-Modal Robotic Platform for Exploration of Titan}

\author{%
Andrea Tagliabue\\ 
Aerospace Controls Lab\\
Dpt. of Aeronautics and Astronautics\\
Massachusetts Institute of Technology\\
Boston, MA 02139\\
(510) 701-4579\\
atagliab@mit.edu\\
\and 
Stephanie Schneider\\
Autonomous Systems Lab\\
Dpt. of Aeronautics and Astronautics\\
Stanford University\\
Stanford, CA 94305-4035\\
(408) 910-5868\\
schneids@stanford.edu\\
\and 
Marco Pavone\\
Autonomous Systems Lab\\
Dpt. of Aeronautics and Astronautics\\
Stanford University\\
Stanford, CA 94305-4035\\
(650) 723 4432\\
pavone@stanford.edu\\
\and 
Ali-akbar Agha-mohammadi\\
Robotic Aerial Mobility Group\\
Jet Propulsion Laboratory\\
California Institute of Technology\\
Pasadena, CA 91109\\
(626) 840-9140\\
aliagha@jpl.nasa.gov\\
}

\maketitle

\thispagestyle{plain}
\pagestyle{plain}

\begin{abstract}
In this paper we present a mission architecture and a robotic platform, the Shapeshifter, that allow multi-domain and  redundant mobility on Saturn's moon Titan, and potentially other bodies with atmospheres. The Shapeshifter is a collection of simple and affordable robotic units, called Cobots, comparable to personal palm-size quadcopters. By attaching and detaching with each other, multiple Cobots can shape-shift into novel structures, capable of
\begin{inparaenum}[(a)]
\item rolling on the surface, to increase the traverse range,
\item flying in a flight array formation, and
\item swimming on or under liquid.
\end{inparaenum}
A ground station complements the robotic platform, hosting science instrumentation and providing power to recharge the batteries of the Cobots. In the first part of this paper we experimentally show the flying, docking and rolling capabilities of a Shapeshifter constituted by two Cobots, presenting ad-hoc control algorithms. We additionally evaluate the energy-efficiency of the rolling-based mobility strategy by deriving an analytic model of the power consumption and by integrating it in a high-fidelity simulation environment. In the second part we tailor our mission architecture to the exploration of Titan. We show that the properties of the Shapeshifter allow the exploration of the possible cryovolcano Sotra Patera, Titan's Mare and canyons. 
\end{abstract}
\vspace{-10pt}
\tableofcontents
\section*{Supplementary material}
\centerline{\href{https://youtu.be/DZ6PLllJFzI}{https://youtu.be/DZ6PLllJFzI}}

\section{Introduction}
\begin{figure}
\centering
\includegraphics[width=\columnwidth]{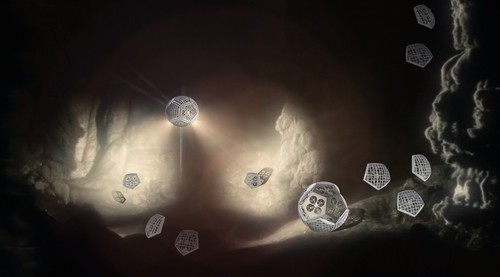}
  	\caption{Artist's representation of the mission concept for the Shapeshifter. A Shapeshifter is constituted by smaller, flying robots that can combine together, morphing into a robot able to roll on the surface or swim in liquid basins. Image credits: Ron Miller, Marilynn Flynn, Jose Mendez.}
\label{fig:intro}
\end{figure}
Titan is arguably the most Earth-like world in the Solar System, and its exploration has stimulated the interest of many scientists and researchers \cite{nasaNIAC}. It has a dense atmosphere, vast dune fields, rain, rivers, and seas that are part of a methane hydrologic cycle analogous to Earth’s water cycle \cite{aharonson2014titan}, \cite{hayes2016lakes}. Titan also has the most complex atmospheric chemistry in the Solar System, that may include prebiotic chemistry similar to that on early Earth before life began \cite{horst2017titan}. There is significant scientific motivation to explore this enigmatic world but two considerations have hampered mission concepts:
\begin{inparaenum}[(a)]
    \item many of the most scientifically enticing locations are difficult to access due to challenging surface conditions such as steep slopes, and 
    \item  a desire to explore all of Titan’s diverse terrains but inability to traverse the long distances between them.
\end{inparaenum} \par Most of the current autonomous mobility systems for planetary exploration are based on monolithic (single agent) robotic platforms, and are usually tailored for the exploration of a specific terrain or domain, such as solid surface (e.g., landers \cite{lebreton2005overview} and rovers \cite{nasaMERS}, \cite{nasaMSL}), liquid basins (e.g., \cite{stofan2013time}), or atmosphere (e.g., \cite{lorenz2018dragonfly}, \cite{lorenz2008review}, \cite{barnes2012aviatr}, \cite{matthies2014titan}). In addition, because of their design philosophy, most platforms only provide redundancy at the component-level and thus do not guarantee margin for failure in high-risk, high-reward exploration scenarios. Platform designs specialized to traverse the challenging environments encountered in different bodies of the Solar System, such as  rough surfaces \cite{elachi2005cassini}, \cite{bibring2005mars}, cliffs \cite{burr2013fluvial}, and subsurface voids \cite{tan2013titan}, incur penalties to the rest of the system design with only a marginal mitigation of the risk. Unknown, unpredictable and constantly-changing \cite{hueso2006methane} environments additionally dictate for the need of flexible, redundant, multi-purpose mobility solutions, which are capable of efficient, across domains locomotion.
\par In this work, we present a multi-agent robotic platform and mission architecture that allow for multi-domain, resilient, and all-access mobility on Titan, and potentially other bodies with atmosphere. The proposed mobility system is based on the concept of shapeshifting, from which the name takes inspiration. The Shapeshifter is a collection of robotic units, called Cobots, capable of attaching and shaping themselves into different forms in order to accomplish various goals, as represented in Figure \ref{fig:intro}. Each Cobot is mechanically simple and economically affordable, comparable to personal palm-size quadcopter, consisting of motor-propeller-based mobility, instrumentation, and electronics. Via an active-controllable magnetic docking mechanism, multiple Cobots can mechanically connect together, morphing into new structures, which include 
\begin{inparaenum}[(a)]
\item a rolling vehicle that rolls on the surface, to increase the traverse range by reducing power consumption,
\item a flight array that can fly and hover above-surface and move in subsurface voids, and
\item a torpedo-like structure to swim under-liquid for chemical and biological measurements.
\end{inparaenum}
The proposed solution is complemented by a lander, called the Home-base, which hosts the power-source necessary to recharge the batteries of the Cobots, the scientific instrumentation and the telecommunication technologies. 
A swarm of Cobots can additionally be used to relocate the Home-base, using collaborative aerial transportation strategies, and to create a communication mesh capable of subterranean exploration and spatio-temporal measurements. 
Due to the inherent redundancy of the multi-agent platform and the low-cost design of each Cobot, high risk tasks can be more easily accepted, as the loss of one or more agents does not compromise any of the functionality of the platform. 
\par The proposed mobility solution is evaluated by showing the flying, docking and rolling capabilities of a simplified Shapeshifter design, constituted by two simplified Cobots. Each Cobot is based on a traditional quadcopter design and is equipped with a hemicylindrical shell, adequate to fly, roll, and withstand small impacts. Additionally, the Cobots are capable of docking and un-docking via \ac{PEM}, allowing the system to shape-shift into a cylindrical structure capable of rolling on different surfaces. By making use of first-principle laws, we derive a model to evaluate the energy efficiency of the pure rolling strategy and compare it with pure-flight strategies. We additionally present an high-fidelity simulation environment of Titan, used to test the proposed motion control algorithms and validate our energy-efficiency analysis. %
\par The remainder of this work is structured as follows: Section \ref{sec:MobilityConcept} describes the proposed platform and details its mobility capabilities, Section \ref{sec:RoboticPlatform} presents a simplified, two-agents prototype of a Shapeshifter and illustrates an energy-efficiency analysis and simulation, Section \ref{sec:MissionArchitecture} lists some key capabilities useful for the future design of the mission architecture, Section \ref{sec:MissionOperationsTitan} proposes a case study for the exploration of Titan, to establish some aspect of mission operations. Conclusion and future works are presented in Section \ref{sec:ConclusionAndFutureWorks}.

\section{Platform and locomotion concepts}
\label{sec:MobilityConcept}
\begin{figure}
\centering
\includegraphics[width=0.95\columnwidth]{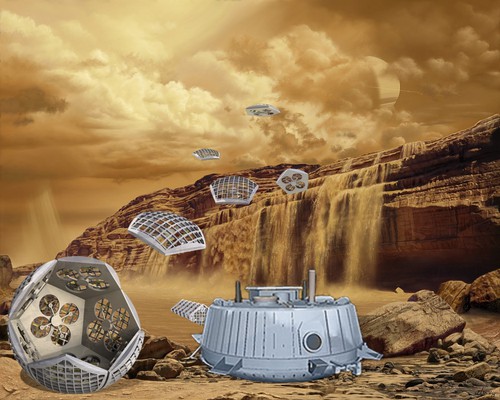}
  	\caption{Artist's representation of the Shapeshifter platform, constituted by Cobots (flying and partially assembled in a sphere, \textit{top and bottom left}) and by a lander called the Home-base \textit{(bottom right)}. Image credits: Ron Miller, Marilynn Flynn, Jose Mendez.}
\label{fig:PlatformOverview}
\end{figure}
In this section, we present the components that constitute the Shapeshifter platform, namely the Cobots and the Home-base ground station, and their key locomotion modes.

\subsection{Platform}
The hardware platform of the Shapeshifter is constituted by two components, the Cobots and the Home-base, as represented in Figure \ref{fig:PlatformOverview}.
\subsubsection{Cobot}
\begin{figure}[h]
    \centering
    \includegraphics[width=0.60\columnwidth]{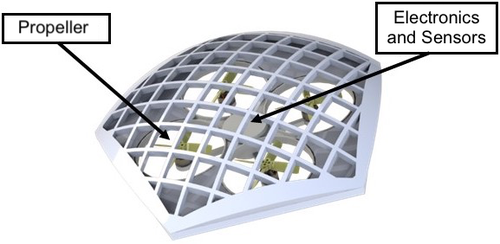}
  	\caption{Artist's representation of a Cobot. }
    \label{fig:Cobot}
\end{figure}
Shapeshifter’s Cobot units are similar to existing, off-the-shelf quadcopters, as represented in Figure \ref{fig:Cobot}. Each Cobot is equipped with four rotating propellers which allow the Cobot to fly. The frame enclosing the Cobot's actuators is equipped with controllable magnets (for example, programmable polymagnets \cite{sullivan2005magnetic}) that allow Cobots to self-assemble and perform shapeshifting. Cobot power is provided by an on-board battery that is recharged at the portable Home-base. Each Cobot is equipped with a camera and Inertial Measurement Unit (IMU), to perform Visual-Inertial based navigation and mapping. Cobots may additionally be equipped with a scoop or tools to collect samples to be analyzed by the Home-base. Their design is complemented by a radio, which allows to communicate with the Home-base and with other Cobots, via a mesh network.

\subsubsection{Home-base}
The design of the Home-base is inspired by the Huygens lander used during the Cassini mission to Titan \cite{liechty2006cassini}.
The main task of the Home-base is to host:
\begin{inparaenum}[(a)]
  \item the instrumentation necessary to perform science measurements and analyze samples collected by the Cobots,
  \item the baseline radioisotope-based power system (RPS), such as an MMRTG \cite{ritz2004multi}, necessary to provide power to the Home-base itself and recharge the batteries of the Cobots, and
  \item to host the equipment necessary to establish a communication link with Earth.
\end{inparaenum}
The Home-base is not equipped with any means of locomotion, but can be collaboratively transported by  a swarm of Cobots. An illustration of the Home-base, based on the Huygens lander design, is depicted in Figure \ref{fig:PlatformOverview}.

\subsection{Locomotion modes}
\begin{figure}
\centering
\includegraphics[width=1\columnwidth]{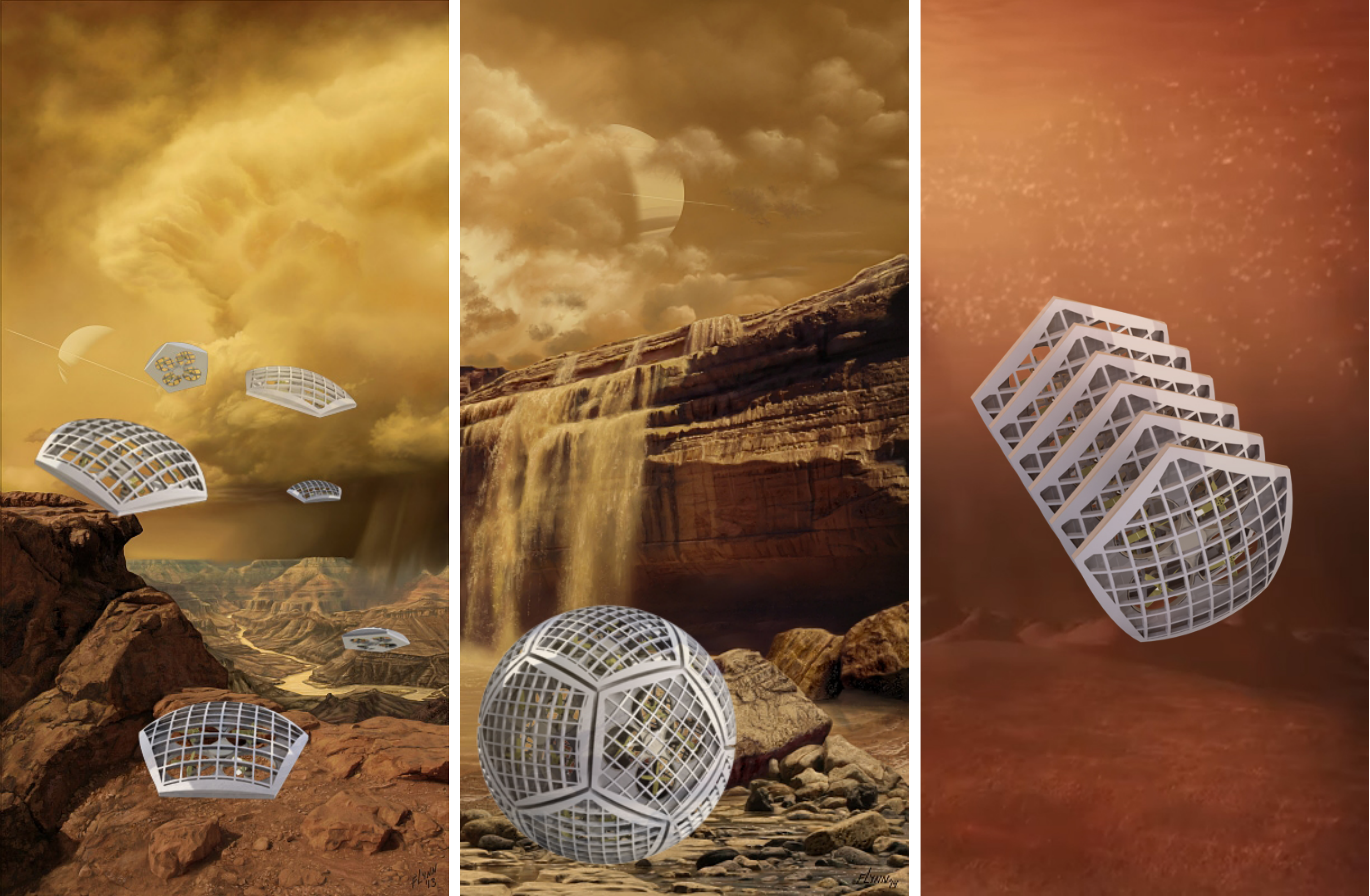}
  	\caption{Artist's concept of the three main locomotion modes of the Shapeshifter (from left to right): flying, rolling on the surface and swimming. Image credits: Ron Miller, Marilynn Flynn, Jose Mendez.}
\label{pic:sys_des:ref_frame}
\end{figure}

In this section, we present the main locomotion modes that the Shapeshifter can adopt by combining multiple Cobots together. An artist's representation of the Shapeshifter in its three main locomotion modes can be found in Figure \ref{pic:sys_des:ref_frame}.

\subsubsection{Flying and flying-array}
Each Cobot can autonomously fly in Titan's atmosphere, for exploration, mapping and sample collecting purpose. A group of Cobots can additionally morph into a flight array, able to to lift and carry heavy objects, such as the portable Home-base, from one mission site to another, or even over lakes and cliffs. %

\subsubsection{Rollocopter}
Shapeshifter is able to morph into a spherical robot, the Rollocopter \cite{rollocopter2019Ieee}, that is able to roll on the surfaces and fly. In our artist's representation shown in Figure \ref{pic:sys_des:ref_frame}, we assume that 12 pentagon-shaped Cobots will morph into a spherical dodecahedron. While rolling, the robot is actuated by the same propellers used for flying. The implications, in terms of control and efficiency, of rolling using the force produced by propellers are studied in our related work \cite{rollocopter2019Ieee}.
\subsubsection{Torpedo}
Each Cobot can autonomously swim underwater or float on the surface, thanks to neutral buoyancy. Multiple Cobots can mechanically dock together in a torpedo-like structure for increased underwater autonomy or propulsion, in case of strong underwater currents. In this paper we mainly focus on the rolling and flying capabilities of the Shapeshifter, and a more detailed description of the \textit{Torpedo} mode is left as future work.

\section{Two-agent robotic platform analysis and implementation}
\label{sec:RoboticPlatform}
In this section, we present a simplified, two-Cobots prototype of the Shapeshifter, which is focused on validating the \textit{Fly mode} and \textit{Rollocopter mode} of the Shapeshifter. 
Despite featuring a reduced number of agents and a simplified mechanical design w.r.t. our original concept, this prototype shows critical feasibility aspects of the mobility of the platform,
such as the ability to create a mechanism for the docking and un-docking of the two agents, and the ability to roll using the thrust produced by propellers. %
We additionally derive a dynamic model of the flying Cobot and rolling Shapeshifter, assuming no slipping. Such a model takes into account the power consumption, with special attention on the aerodynamic power, and it is used to compare the energy required to fly and roll on different terrains (in terms of friction and slopes) on Titan. We present a simple motion control algorithm developed to achieve the rolling behavior, which is validated, together with the energy-efficiency analysis, on a high-fidelity simulation environment of Titan. Last, we show preliminary experiments that validate our mechanical design and motion strategy, showing a docking, rolling and un-docking maneuver.

\subsection{Mechanical design}
\begin{figure}
\centering
\includegraphics[width=1\columnwidth]{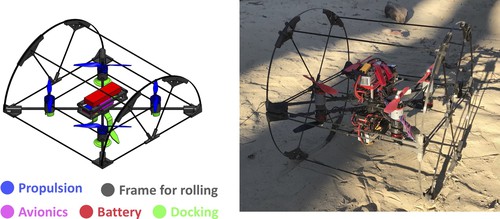}
      \caption{\textit{Left:} Sub-components of a Cobot prototype. \textit{Right:} Our physical prototypes of two Cobots docked together forming a Rollocopter, capable of rolling on sand for increased range w.r.t. fly.}
\label{fig:PlatformImplementationOverview}
\end{figure}
A mechanical design of the two-agents Shapeshifter is based on Cobots equipped with a hemispherical shell so that the docking of two agents creates a cylindrical structure adequate for rolling in one dimension. The prototype of the platform is represented in Figure \ref{fig:PlatformImplementationOverview}, where we have highlighted the different sub-components that constitute the Cobot. The two Cobots are identical, with the exception of the docking mechanism and the spin direction of the propellers.

\subsection{Dynamic models for rolling and flying robots}\label{sec:dynamics}
In this section, we derive the dynamic model of a rolling Shapeshifter (Rollocopter), assuming that rolling happens without slipping, and we describe the dynamic model of a flying Cobot. We additionally outline an induced-velocity model for power consumption as a function of the rotor thrust (see e.g. \cite{johnson2012helicopter}, \cite{tagliabue2019model}) both for the flying and rolling cases. 

\subsubsection{Reference frame}
\begin{figure}
\centering
\includegraphics[width=0.9\columnwidth]{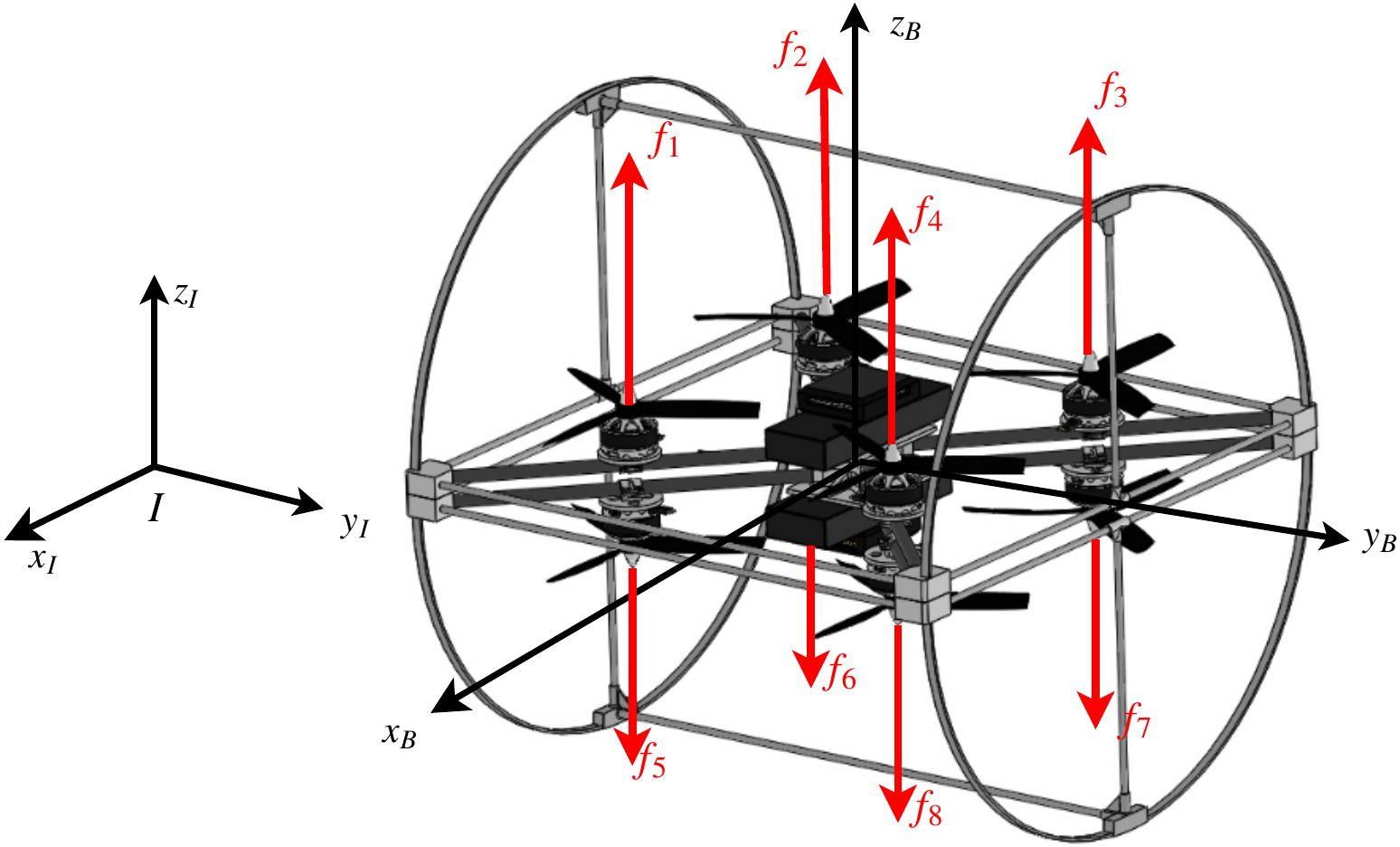}
      \caption{Model of the two-agents Shapeshifter, where we have highlighted the inertial reference frame $I=(x_I, y_I, z_I)$, the body-fixed reference frame $B=(x_B, y_B, z_B)$ and the forces produced by the actuators.}
\label{fig:3DShapeshifterFramesAndActuaturs}
\end{figure}

For both the rolling and flying vehicles, we define an inertial reference frame $I=(x_I, y_I, z_I)$ and a non-inertial reference frame $B=(x_B, y_B, z_B)$ fixed in the \ac{CoM} of each vehicle. The frames, in the case of the rolling robot, are represented in Figure \ref{fig:3DShapeshifterFramesAndActuaturs}.

\subsubsection{Notation declaration} The notation $\prescript{}{W}{\boldsymbol{r}} = \prescript{}{W}{(r_x, r_y, r_z)}$ denotes a vector defined in the Cartesian reference frame $W$. The matrix $\prescript{}{A}{\boldsymbol{R}}_{BC}$ denotes the rotation matrix from the coordinate frame $C$ to $B$, defined in $A$.

\subsubsection{Rolling Shapeshifter}
The dynamic equations of the rolling Shapeshifter are derived by using the Newton-Euler method: 
\begin{align}
\prescript{}{I}{\ddot{\boldsymbol{x}}} = & \frac{1}{m} \big ( \prescript{}{I}{\boldsymbol{R}}_{IB}\prescript{}{B}{\boldsymbol{f}_\text{cmd}} - m\prescript{}{I}{\boldsymbol{g}} + \prescript{}{I}{\boldsymbol{r}} + \prescript{}{I}{\boldsymbol{f}_\text{drag}}\big)
\label{eq:RollocopterDynamics:translation}
\end{align}
\begin{equation}
\begin{split}
\prescript{}{B}{\dot{\boldsymbol{\omega}}} =  \boldsymbol{J}^{-1} \big ( & \prescript{}{B}{\boldsymbol{\tau}_\text{cmd}} - \prescript{}{B}{\boldsymbol{\omega}} \times \boldsymbol{J} \prescript{}{B}{\boldsymbol{\omega}} \\ &
- l\prescript{}{I}{\boldsymbol{R}}_{IB}^{-1} (\prescript{}{I}{\boldsymbol{n}} \times \prescript{}{I}{\boldsymbol{r}}) + \prescript{}{B}{\boldsymbol{\tau}_\text{rolling}} \big ) 
\label{eq:RollocopterDynamics:rotational}
\end{split}
\end{equation}
where $m$ and $\boldsymbol{J}$ represent, respectively, the mass and inertia of the vehicle; the vectors $\boldsymbol{x}, \dot{\boldsymbol{x}}, \ddot{\boldsymbol{x}}$ represent the position of the robot and its derivatives, while $\boldsymbol{\omega}, \dot{\boldsymbol{\omega}} $ the angular rates and their derivatives. The attitude is represented by the rotation matrix $\prescript{}{I}{\boldsymbol{R}}_{IB}$, defining a rotation from $B$ to $I$, while $\prescript{}{I}{\boldsymbol{n}}$ and $\prescript{}{I}{\boldsymbol{r}}$ represent, respectively, the normal of the plane on which the vehicle is rolling and the reaction force with such a plane, expressed in $I$. The forces and torques produced by the actuators are defined by $\prescript{}{B}{\boldsymbol{f}}_\text{cmd} = (0, 0, f_\text{cmd})$, and $\prescript{}{B}{\boldsymbol{\tau}}_\text{cmd}$.  The total thrust force $f_\text{cmd}$ is defined as: 
\begin{align}
    f_\text{cmd} = \sum_{i=1}^{4}f_i - \sum_{i=5}^{8}f_i %
\end{align} 
under the assumption that all the propellers are placed on parallel planes. The propellers are placed so that opposite or adjacent pairs spin in opposite directions. For example, following the same numbering scheme adopted in Figure \ref{fig:3DShapeshifterFramesAndActuaturs}, the pair of propellers (1, 5) have opposite directions of rotation, as well as the pair (1, 2). 
For simplicity, we assume that the vehicle rolls without slipping around $y_B$.
The torque due to the deformations caused by the interaction between the vehicle and the terrain 
is thus modeled as
\begin{align}
    \prescript{}{B}{\boldsymbol{\tau}}_\text{rolling} = (0, C_\text{rr} \prescript{}{I}{\boldsymbol{r}} \cdot \prescript{}{I}{\boldsymbol{n}} l, 0) 
    \label{eq:rolling}
\end{align}
where $\cdot$ represents the scalar product, $C_\text{rr}$ the rolling resistance coefficient and  $l$ the radius of the cylindrical shell of the robot.
We additionally assume that the aerodynamic drag force $\boldsymbol{f}_\text{drag}$, applied in the \ac{CoM} of the robot, is function of the second power of the velocity of the vehicle:
\begin{align}
    \prescript{}{I}{\boldsymbol{f}}_\text{drag} = -\frac{1}{2}C_\text{d} \rho A ||\prescript{}{I}{\dot{\boldsymbol{x}}}||\prescript{}{I}{\dot{\boldsymbol{x}}}
    \label{eq:drag}
\end{align}
where $C_\text{d}$ is the drag coefficient, and $A$ is the aerodynamic area, computed as the area of a Cobot's rectangular base projected on the plane orthogonal to the velocity of the vehicle $\prescript{}{I}{\dot{\boldsymbol{x}}}$. We observe that scalar $A$ is a function of the attitude of the robot.

\subsubsection{Flying Cobot}
The dynamic model of a flying Cobot is a special case of the rolling Shapeshifter, and can be obtained by Equations (\ref{eq:RollocopterDynamics:translation}) and (\ref{eq:RollocopterDynamics:rotational}) assuming $\boldsymbol{r} = \boldsymbol{0}$ and, as a consequence, $\boldsymbol{\tau}_\text{rolling} = \boldsymbol{0}$. %

\subsection{Power consumption model for rolling and flying robots}
We assume that the total power consumption of the robot is directly proportional to the aerodynamic power produced by the propellers, according to the commanded thrust and the motion of the robot, as shown in \cite{tagliabue2019model}, \cite{ware2016analysis}. We disregard the power consumption of other processes (e.g. thermal) in our analysis, as we don't expect many discrepancies between the two mobility modes.
Our induced velocity model relates individual rotor thrust to aerodynamic power, as detailed in \cite{leishman2006principles}, assuming forward-flight regime. %
The forward flight power-thrust model for the $i\textit{-th}$ propeller can be expressed as:
\begin{align}
    P_i = \frac{f_i(\nu_i - \nu_\infty \sin \alpha)}{\eta_p \eta_m \eta_c}
\end{align}

Power consumption of rotor $i$, $P_i$, is defined as a function of rotor thrust ($f_i$), induced velocity ($\nu_i$), freestream airspeed ($v_\infty$), and angle of attack ($\alpha$). For this analysis, we assumed a propeller efficiency of $\eta_p$ = 0.6, motor efficiency of $\eta_m$ = 0.85, and controller efficiency $\eta_c$ = 0.95. 
We assume that this forward-flight model applies to the rolling configuration as well, justified by the observation that all rotors in the Rollocopter are moving forward relative to freestream at all times during a rotation. We neglect any effect that potential vortex ring states might have on total power consumption.

\subsection{Motion control strategies for rolling and flying robots}
\subsubsection{Rolling Shapeshifter}
\label{sec:rolling_control} 
For this initial study, we implemented a simple, centralized control strategy that tracks a desired body rate $\prescript{}{B}{\boldsymbol{\omega}}_\text{des}$ by applying a pure torque on the Rollocopter, i.e., producing zero net thrust. A detailed derivation for the control strategy of a conceptually similar platform is proposed in our related work \cite{rollocopter2019Ieee}. The desired torques $\prescript{}{B}{\boldsymbol{\tau}}_\text{cmd}$, expressed in the frame $B$ of the rolling vehicle, are computed according to a proportional-integral (PI) controller, using the measurements $\prescript{}{B}{\hat{\boldsymbol{\omega}}}$ of the body angular rates provided by the IMU on-board one of the two docked Cobots:
\vspace{-2pt}
\begin{align}
    &\boldsymbol{e} = \prescript{}{B}{\boldsymbol{\omega}}_\text{des} - \prescript{}{B}{\hat{\boldsymbol{\omega}}} \\
    \prescript{}{B}{\boldsymbol{\tau}}_\text{cmd} & = \boldsymbol{K}_\text{p}\boldsymbol{e} + \boldsymbol{K}_\text{i}\int_{t_0}^{t} \boldsymbol{e} dt
\end{align}\vspace{-2pt}
where $\boldsymbol{K}_\text{p}$ and $\boldsymbol{K}_\text{i}$ are diagonal matrices, tuning parameters of the controller. 
Given the commanded torque $\prescript{}{B}{\boldsymbol{\tau}}_\text{cmd}$, a rotation speed input $n_i$ for the $i\textit{-th}$ propeller, with $i = 1, ..., 8$ is produced in the following way.
First, we define the quadruple $(f_\text{A}, ..., f_\text{D})$, where every entry corresponds to the force produced by each pair of opposite propellers (e.g. $f_\text{A} = f_\text{1} - f_\text{5}$, $f_\text{B} = f_\text{2} - f_\text{6}$, etc).
Second, we derive an expression which relates the forces $f_\text{A}, ..., f_\text{D}$ with
\begin{inparaenum}[(a)]
 \item the total torques produced by the propellers, which we assume to be equivalent to $\prescript{}{B}{\boldsymbol{\tau}}_\text{cmd}$, and
 \item the force $f_\text{cmd}$ produced along the positive $z$ axis of $B$, and expressed in body $B$ frame.
\end{inparaenum} 
Such expression is defined as:
\begin{equation}
    \begin{bmatrix}
    f_\text{cmd} \\ \boldsymbol{\tau}_\text{cmd} \\
    \end{bmatrix} =
    \boldsymbol{M} 
    \begin{bmatrix}
        f_\text{A} \\ f_\text{B} \\ f_\text{C} \\ f_\text{D} \\
    \end{bmatrix}
    \label{eq:AllocationStrategy}
\end{equation}
where $\boldsymbol{M}$ is a squared, full-rank matrix by the design of the system, and corresponds to the inverse of the allocation matrix. The matrix $\boldsymbol{M}$ is defined as: 
\begin{align} 
\boldsymbol{M} = 
    \begin{bmatrix}
    1 & 1 & 1 & 1 \\
    -c & c & c & -c \\
    -c & -c & c & c \\
    -k_\tau & k_\tau & -k_\tau & k_\tau \\
    \end{bmatrix}
\end{align}
with $c = \frac{a}{\sqrt{2}}$, where $a$ is the arm length of each propeller to the \ac{CoM}, and $k_\tau$ is a constant with maps the $i\textit{-th}$ propeller's thrust $f_i$ to the aerodynamic torque $\tau_i$, according to $\tau_i = k_\tau f_i$.
Last, because we have assumed that the net force produced by propellers has to be zero, we impose $f_\text{cmd} = 0$ and solve Equation \ref{eq:AllocationStrategy} by inverting the square matrix $\boldsymbol{M}$, 
allowing us to find the desired force that has to be produced by each propeller pair.
The desired rotation speed to be commanded to the propellers in each propeller pair can be retrieved in a straightforward way, for example, in the case of the propeller pair A (constituted by propeller 1 and 5):
\begin{equation}
    (n_1, n_5) 
    = 
    \begin{cases}
        (\sqrt{f_\text{A}/k_\text{t}}, 0) & \mbox{if } f_\text{A} \geq 0 \\
        (0, \sqrt{-f_\text{A}/k_\text{t}}) & \mbox{if } f_\text{A} < 0 
    \end{cases}
\end{equation}
where $k_\text{t}$ relates the force produced by each propeller with its rotation speed according to $f_i = k_\text{t} n_i^2$.

\subsubsection{Flying Cobot}
For the flying Cobot, we employ a common cascaded control architecture \cite{Controll67:online} provided by the embedded microcontroller framework \cite{meier2015px4}.

\subsection{Framework for the energy efficiency analysis}

In this part, we introduce the approach used for the energy analysis.
We focus on the steady-state, planar motion of the robots ($\ddot{\boldsymbol{x}} = \boldsymbol{0},  \dot{\boldsymbol{\omega}} = \boldsymbol{0}$), and assume that the motors are controlled to apply pure torque (zero net thrust), implementing the presented control strategy. In the following analyses, $\theta$ defines the slope of the terrain, and $\alpha$ defines the angle between the body frame and the inertial coordinate system, as rotated about $\hat{y}_I = \hat{y}_B$ (\textit{pitch} angle). Cobot dimensions and other parameters used for the analysis are defined in Table \ref{table:dimensions}.

\newcommand\Tstrut{\rule{0pt}{2.6ex}}       %
\newcommand\Bstrut{\rule[-0.9ex]{0pt}{0pt}} %
\newcommand{\TBstrut}{\Tstrut\Bstrut} %

\begin{table}[h]
\begin{center}
 \begin{tabular}{| r | l | l |} 
 \hline
 \textit{Symbol} &  \textit{Meaning} & \textit{Value} \TBstrut\\
 \hline \hline
 $h$ & \textit{rolling}: height from one rotor & 0.16 m \Tstrut\\
  & to opposite rotor & \Bstrut\\
  & \textit{flying}: height from rotor to base & 0.08 m \TBstrut\\
 \hline 
 $l$ & radius of cylinder & 0.2 m \TBstrut\\
 \hline
 $w$ & width of cylinder & 0.4 m \TBstrut\\
 \hline
 $C_\text{d}$ & drag coefficient & 2.1 \TBstrut\\
 \hline 
 $C_\text{rr}$ & rolling resistance coefficient & 0.01 - 0.2 \TBstrut\\
 \hline 
 \hline
\end{tabular}
\end{center}
\caption{Parameters for the two-Cobot Shapeshifter.}
\label{table:dimensions}
\end{table}

\subsubsection{Rolling Shapeshifter}
\begin{figure}[h]
    \centering
    \includegraphics[width=\linewidth]{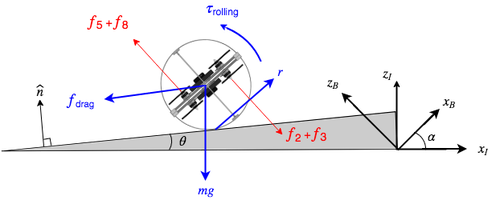}
    \caption{Free body diagram: rolling uphill. Thrust from four rotors produces a pure torque. Rollocopter is propelled by the ground reaction force (assume rolling without slipping) up a slope defined by $\theta$.}
    \label{fig:rolling_fbd}
\end{figure}
The free-body diagram used in the planar-motion analysis of the rolling configuration is shown in Figure \ref{fig:rolling_fbd}. Torque due to rolling resistance is calculated as per Eq. (\ref{eq:rolling}). This analysis assumes a rolling friction coefficient between that of consolidated soil ($C_\text{rr} = 0.01$) and tires on loose sand ($C_\text{rr} = 0.2$), consistent with data gathered from the Huygens Surface Science package \cite{zarnecki2005soft}.
Drag force on the Rollocopter is calculated based on Eq. (\ref{eq:drag}), where drag coefficient, $C_\text{d} = 2.1$, is approximated based on Titan's atmosphere \cite{liechty2006cassini}. The aerodynamic area, $A$, is computed as the area of a Cobot's rectangular base projected onto the plane orthogonal to its velocity vector, as a function of the Rollocopter's orientation (\ref{eq:Ax}). 

\begin{equation}
    A = (h \mid\cos\alpha\mid + 2l\mid\sin\alpha\mid)w
    \label{eq:Ax}
\end{equation}

\subsubsection{Flying Cobot}
\begin{figure}[h]
    \centering
    \includegraphics[width=0.8\linewidth]{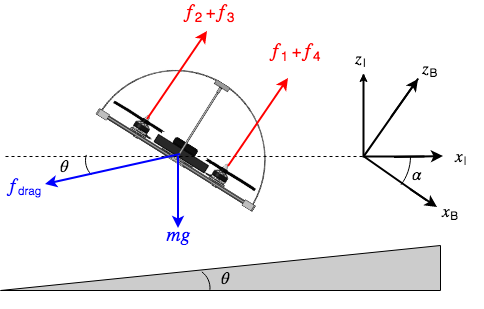}
    \caption{Free body diagram: flying. Cobot flies at a constant height above a hill defined by $\theta$. Angle of attack $\alpha$ is determined by the angle required to keep the Cobot at a constant altitude.}
    \label{fig:flying_fbd}
\end{figure}
Figure \ref{fig:flying_fbd} shows a free-body diagram for the flying configuration of a single Cobot. In this configuration, $\alpha$ is determined based on the angle of attack required for the Cobot to fly at a constant height above the ground. Aerodynamic drag force is again given by (\ref{eq:drag}), and aerodynamic area by (\ref{eq:Ax}).

\subsection{Results for the two-Cobot Shapeshifter}
In this section, we present the preliminary results for the mobility validation and energy-efficiency analysis of the two-agents Shapeshifter. We start by introducing a high-fidelity simulation environment of Titan, developed to test our control algorithms, validate our energy-efficiency analysis, and provide a way to simulate mobility aspects of the proposed mission. We then present a physical prototype of the platform, and use it to validate the flying, docking, un-docking, and rolling capability of the proposed design. We conclude this section by showing the results of our energy-efficiency analysis. The results highlight the mechanical feasibility of having a platform capable of both flying and rolling on a sandy terrain, as well as the energy savings realized by morphing into a rolling vehicle.

\subsubsection{Simulation environment}
\begin{figure}
    \centering
    \includegraphics[clip,width=\columnwidth]{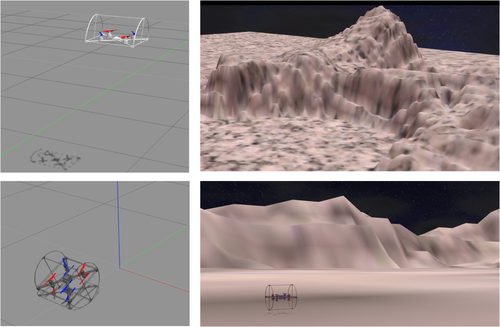}
    \caption{High-fidelity simulation based on ROS/Gazebo of the Shapeshifter on Titan. \textit{Top left:} simulated model of a Cobot. \textit{Top right:} model of the Sotra-Patera region, obtained by elevation maps of Titan reconstructed from images captured by the Cassini mission. \textit{Bottom right:} the simulated Shapeshifter (assembled as Rollocopter) near Sotra-Patera on Titan. \textit{Bottom left:} the simulated model of the Shapeshifter assembled as Rollocopter.}
    \label{fig:gazebo_titan}
\end{figure}

One goal of this work is to construct a physics-based simulation to verify the analytical energy analysis, as well as provide a tool for further mobility analyses and development of control strategies. We chose to use Gazebo \cite{koenig2004design} as the simulation environment because of its realistic physics engine, including aerodynamic drag and rolling resistance, as well as existing quadrotor packages \cite{Furrer2016}. With Gazebo, the user can place Cobots in either flying or two-agents rolling configurations, then input waypoints for the agent; the simulation outputs power and energy data in real-time, providing a versatile platform for testing different Shapeshifter missions. \newline Our simulation setup is not limited to simulate the robot, but includes a 3D model of the Sotra Patera area on Titan. This model has been obtained via an elevation map computed from images captured by the Cassini mission. Figure \ref{fig:gazebo_titan} \textit{(bottom right)} shows the Shapeshifter rolling on a surface generated from a depth elevation map of Sotra Patera on Titan \textit{(top right)}. We also consider a basic simulation with one flying Cobot and one Rollocopter traversing a flat surface, shown in Figure \ref{fig:gazebo_titan} \textit{(top and bottom left)}, to isolate the mobility primitives.

\subsubsection{Hardware implementation details}
In this part, we present details of the the hardware prototype (two-Cobot Shapeshifter) built to validate the flying and rolling mobility modes of the Shapeshifter.
Each Cobot consists of four 6-inch propellers, \textit{EMax 2300 kV} brushless-DC motors, and a three-cell 2200 mAh battery, which provides approximately eight minutes of flight (on hover, on Earth). The side length and the diameter of the cylinder when two Cobots are docked is 0.4 m; the total weight of each Cobot is approximately 0.8 kg. Enough payload transportation capacity is guaranteed by the maximum thrust produced by the propellers, which corresponds to approximately 32 N. On-board computing power and IMU are provided by a Pixwhawk-mini running a PX4 flight stack \cite{meier2015px4}. The two Cobots are identical, with the exception of the position of the magnets and mechanical funnels for the docking mechanism.
\textbf{Shell for rolling and fly:} Each Cobot is equipped with a shell adequate for flying and capable of withstanding small impacts during rolling. The shell is designed using carbon fiber tubes, adequate for this task because they are stiff and lightweight. The carbon fiber tubes are connected together via 3D printed joints. 
\textbf{Docking mechanism:}
\begin{figure}
\centering
\includegraphics[width=1\columnwidth]{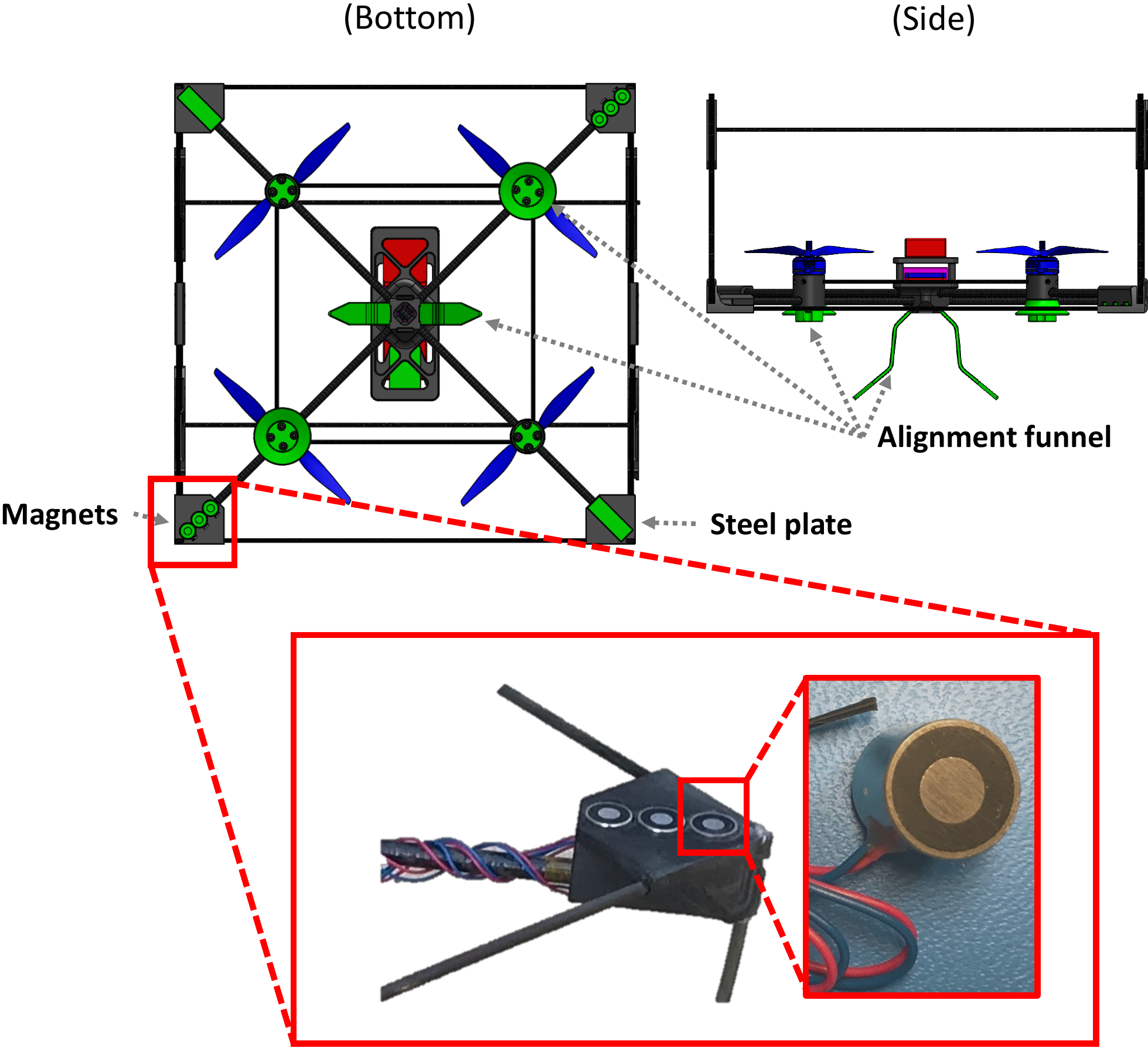}
  	\caption{Illustration of the magnetic and mechanical docking mechanism used to connect two Cobots.}
\label{fig:DockingMechanism}
\end{figure}
The docking mechanism is based on 12 permanent electromagnets (PEMs) mounted at two diametrically opposite extremities of each Cobot. Each PEM weighs approximately 10 g and produces a normal force of 15 N when connected to a 2 mm thick plate of steel, and when not powered. When powered, the magnet produces approximately 0 N of force. Thanks to this property, in order to maintain the two agents docked, no power is required. The PEMs can be activated and deactivated by a micro-controller interfaced with the on-board computer.
The docking mechanism is additionally constituted by mechanical funnels connected at the bottom of each Cobots, used to compensate for misalignments during the docking phase and to cancel the shear forces between the agents, during rolling. A representation of the docking mechanism, where its main components have been highlighted, can be found in Figure \ref{fig:DockingMechanism}.

\subsubsection{Mobility validation experiments}
\begin{figure*}
    \centering
    \includegraphics[width=0.86\linewidth]{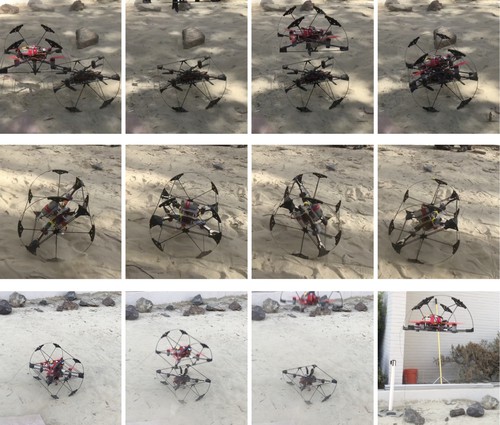}
    \caption{Frames from the video-clip (see Supplementary Material) of the experiments conducted with our two-Cobot Shapeshifter. \textit{Top:} Docking sequence. \textit{Center:} Rolling sequence. \textit{Bottom:} Un-docking sequence.}
    \label{fig:shapeshifter_movie}
\end{figure*}
In this part, we present the experimental results of the docking, un-docking and rolling maneuver obtained with our prototype, which are represented in Figure \ref{fig:shapeshifter_movie}. 
\textbf{Docking}: As represented in the fist row of Figure \ref{fig:shapeshifter_movie}, a pilot remotely controls the flying Cobot while the second agent is on the ground, with the propellers pointed towards the ground. This experiment shows that docking is possible despite the limited flying accuracy of a human pilots, as a result of the alignment funnels and the strong magnetic field created by the magnets. We verify that docking has successfully happened by manually lifting one of the Cobots. \textbf{Rolling}: Once docked, all the motors of the agents are manually connected to the on-board controller of one robot. This limitation will be overcome in future works, for example by establishing a wireless link between the Cobots or by developing a decentralized control strategy. In our experiment, shown in the second row of Figure \ref{fig:shapeshifter_movie}, a pilot controls the rolling motion via a remote control (RC), configured to send the desired angular rates $\boldsymbol{\omega}_\text{cmd}$. Preliminary experiments show that the vehicle can effectively roll on a sandy terrain, uphill and in small dunes. The cylindrical design, anyway, severely limits the yawing (turning) capabilities of the robot. 
\textbf{Un-docking:} The \acp{PEM} have been configured so that they can be remotely turned off. From the RC, two pilots simultaneously disable the \acp{PEM} while one of the Cobots takes off. The experiment is shown in the third row of Figure \ref{fig:shapeshifter_movie}.

\subsubsection{Energy analysis}
\begin{figure}
    \centering
    \includegraphics[width=\linewidth]{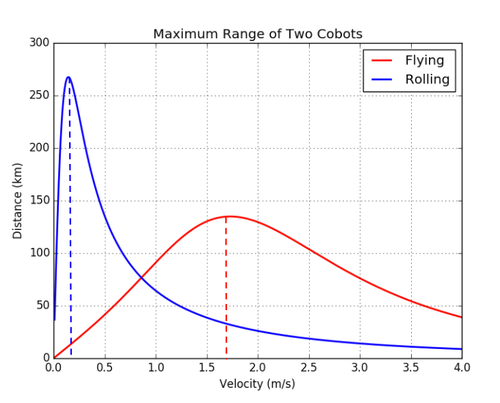}
    \caption{Range vs. velocity for two Cobots on Titan. Maximum achievable range for flying and rolling along a surface of consolidated soil. Dashed lines indicate optimal velocity for each configuration.}
    \label{fig:range_v_velocity}
\end{figure}

In this section, we aim to use our dynamic model to determine environmental conditions for which rolling is the more energy-efficient mobility primitive, as well as the conditions for which it is more efficient to fly. To make this distinction, we focus on developing a functional relationship between terrain primitives and the required energy of mobility. Since the proposed mission involves traversal over long distances of Titan's surface, the main objective of this analysis is to determine the maximum expected steady-state range of the Shapeshifter, both for rolling and flying.
The following results employ the analytical model and pure torque control strategy, 
applying physical parameters for Titan: $g$ = 1.352 m/s$^2$ and $\rho$ = 5.4 kg/m$^3$. To get numerical results, we assume each Cobot has a 870 kJ battery (2200mAh at 11V). 
Figure \ref{fig:range_v_velocity} shows the maximum achievable range for both flying and rolling along a surface of consolidated soil ($C_{rr} = 0.01$); the dashed line indicates the velocity that maximizes range for that configuration. At an optimal velocity of 0.14 m/s, two rolling Cobots are able to travel 267 km, while the maximum range for two flying Cobots is only 135 km, traveling at their optimal velocity of 1.7 m/s.

\begin{figure}[h]
    \centering
    \includegraphics[width=\linewidth]{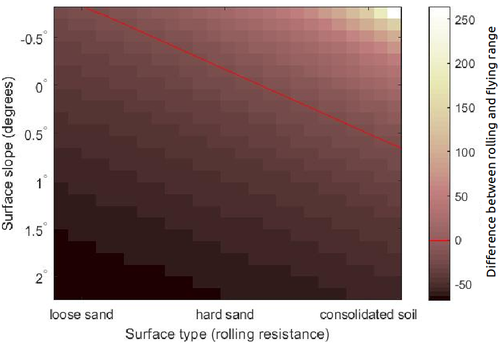}    \caption{Advantage of rolling vs. flying, as it relates to surface slope and rolling resistance. The heat-map represents the difference in range (expressed in kilometers) between flying and rolling (for example, rolling on consolidated soil with $\approx - 0.5^\circ$ of surface slope guarantees $\approx 200 km$ of range more than flying). Flying range is on the order of 130km for all terrains; add flying range to results shown here to get total rolling range.}
    \label{fig:multi_dim}
\end{figure}

As expected, rolling is more efficient than flying over ideal surface conditions. To optimize energy usage while traversing a non-ideal region, we must develop a relationship between terrain characteristics and power required for traversal. Figure \ref{fig:multi_dim} considers terrains that vary in surface traction from consolidated soil to loose sand, and in steepness from -0.5$^\circ$ to +2$^\circ$. Such steepness range is chosen because contains the transition line in which flying becomes more efficient than rolling. For each surface type and mobility method, we compute maximum range assuming the agents travel at their corresponding optimal velocities. By considering the difference in achievable range for each configuration, we can see where rolling is favorable (above red line) vs. flying (below line).
These results demonstrate that neither rolling nor flying consistently outperforms the other; rather, each configuration optimizes energy efficiency for a different set of conditions.
Especially since the characteristics of Titan's surface are largely unknown, a shape-shifting platform is crucial to accommodate unexpected surface conditions.
Furthermore, this relationship between terrain and mobility can be used to build a traversability map, and ultimately plan energy-efficient trajectories that optimize the Shapeshifter's route as well as mobility configuration, to take full advantage of the multi-modal architecture.

\section{Shapeshifter capabilities}
\label{sec:MissionArchitecture}

\begin{figure}
\centering
\includegraphics[width=1\columnwidth]{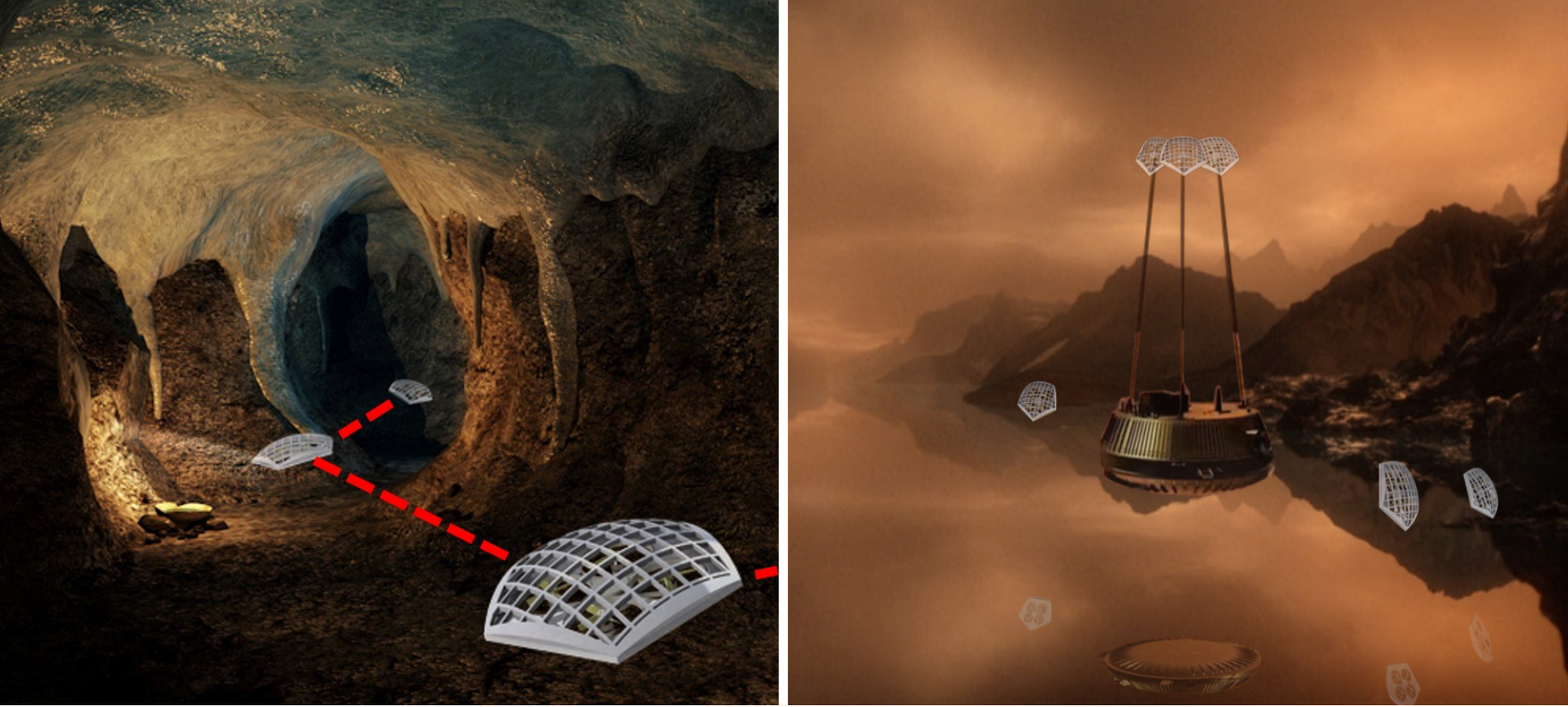}
  	\caption{Artist's representation of (\textit{left}) the Shapeshifter  exploring subterranean environments, maintaining a communication link with the surface, and (\textit{right}) Shapeshifter carrying the Home-base using the Flying-array configuration. Image credits: Ron Miller, Marilynn Flynn, Jose Mendez.}
\label{fig:ExploringCaveAndCarryHomeBase}
\end{figure}

In this section, we present some of the key operation capabilities  of the Shapeshifter platform. Such capabilities leverage the multi-agent nature of the system, as well as the different locomotion modes available and the possibilities of interaction with the Home-base; they constitute building blocks for the design of a mission on Titan using the Shapeshifter platform. The capabilities are grouped according to the three main locomotion configurations: Flying, Rollocopter and Torpedo. The estimates for the range and autonomy of the platform are based on the analysis presented in Section \ref{sec:RoboticPlatform}.
\subsection{Flying and flying-array capabilities}

\begin{itemize}
    \item \textbf{Map large areas}: A group of flying Cobots can search for desired scientific objectives and create a detailed topographical map of Titan.  This allows the Shapeshifter to probe different kinds of surfaces, including very rough terrains and cliffs.
    According to the energy-efficiency analysis presented in Section \ref{sec:RoboticPlatform}, with a maximum range of 130 km, a team of Cobots flying radially outward from the Home-base could survey a circular area of over 10,000 km$^2$, and return home safely to recharge.
    \item \textbf{Create a communication mesh}: To maintain communication from under the surface (e.g., while exploring a cave), Shapeshifter can disassemble into Cobots that will maintain a network of communication nodes and line-of-sight from, e.g., a deep cryovolcanic lava tube to Titan’s surface. An artist's representation of this functionality is included in Figure \ref{fig:ExploringCaveAndCarryHomeBase}.
    \item \textbf{Collect spatio-temporal measurements}: By distributing multiple Cobots in a wide area, observations of a phenomena can be correlated not only with respect to time, but also space. This can be useful to study the evolution of the storms present on Titan \cite{nasaTitanStorms}. 
    \item \textbf{Collect samples}: Cobots can collect samples of rocks and terrain using an on-board scoop, whose design is left as future work. The collected samples are then analyzed by the Home-base. With approximately 30 N of maximum thrust available, a single Cobot could carry more than 20 kg of samples back to the Home-base. Maximum payload transportation capacity, anyway, could be reduced due to the aerodynamic drag of the payload (i.e. due to increased atmospheric density w.r.t. Earth).
    \item \textbf{Transport the Home-base}: Shapeshifter can morph into a flight array of Cobots to lift and carry the portable Home-base from one mission site to another, as represented in an artist's concept in Figure \ref{fig:ExploringCaveAndCarryHomeBase}, by employing a decentralized collaborative aerial transportation strategy such as \cite{collaborativeTransportation}. The paylaod capacity of the twelve Cobots envisioned to be part of the mission can  easily reach more than 200 kg on Titan, sufficient to carry the Home-base. 
\end{itemize}
\subsection{Rollocopter capabilities}
\begin{itemize}
    \item \textbf{Traverse long distances}:  As shown in our analysis, morphing into a sphere can be a more efficient way of traversing long distances. By taking advantage of energy-efficient mobility, the Shapeshifter can increase its range on a single charge to up to 260 km, approximately doubling the range in flight mode. This corresponds to a reachable area of more than 50,000 km$^2$, radially outward from the stationary Home-base. %
    \item \textbf{Explore caves}: Shapeshifter can explore subsurface voids, including cryovolcanic and karstic caves. For narrow passages, the Rollocopter configuration allows resilience to collisions. It allows to bounce off of walls to go through cracks, holes, and narrow passages to reach science targets.
\end{itemize}
\subsection{Torpedo capabilities}
\begin{itemize}
    \item \textbf{Above and sub-surface navigation}: The propeller-based propulsion of each Cobot can generate thrust in gas and liquid environments \cite{diez2017unmanned}. Thus, Shapeshifter can in principle navigate and explore above and below Titan’s mare, such as the Ligeia Mare represented in Figure  \ref{fig:LigeiaMare}. These functionalities can be further developed by leveraging studies on motors and propulsion system for Titan’s hydrocarbon lakes \cite{hartwig2016exploring}. A more detailed analysis of the Torpedo mode of the Shapeshifter is left as future work.
\end{itemize}

\subsection{Autonomy}
One of the main challenges for the proposed mission is autonomy, including localization, 3D mapping, obstacle avoidance, self-assembly and risk-aware decision making. In this section, we propose approaches that can be of interest for the Shapeshifter.
\subsubsection{State estimation, localization and mapping}
For navigation, we expect to rely on resource-constrained VIO (Visual Inertial Odometry) \cite{li2012vision}, which allows to achieve a cm-level localization accuracy when coupled with state-of-the-art localization solutions (e.g. via Distributed Pose-Graph Optimization), using on-board cameras and computing unit such as \cite{Snapdrag52:online}. In mapping mission scenarios, we envision the ability to create precise 3D maps with limited onboard computation power (following, for example, \cite{mu2015two}, \cite{agha2017confidence}, \cite{lajoie2019door}), in order to be able to compress the large number of images acquired by each Cobot into a representative map that can be easily shared with the Home-base. Potential perception challenges in (methane) fog or under-liquid \cite{garcia2017exploring} \cite{maldonado2016robotic} will be studied.
\subsubsection{Motion planning and control}
We use sampling-based methods \cite{janson2015fast}, \cite{pavone2007decentralized} to plan precise motions in 3D to accomplish shapeshifting behaviors, and adaptive trajectory-generation strategies such as \cite{tagliabue2019model} to guarantee that the Shapeshifter always operates at the most energy-efficient velocity. Distributed algorithms are employed in the case the Cobots have to find each other when lost or have to transport the Home-base without relying on any communication network \cite{collaborativeTransportation}, \cite{tagliabue2017collaborative} for example due to strong electromagnetic interferences. 
\subsubsection{Decision making}
A fundamental aspect for the Shapeshifter is the ability to efficiently make complex decisions under uncertainty. In order to operate at its full capabilities, for example, the Shapeshifter will face the constant need to decide in which configuration to morph, based on multiple factors such as the environment (e.g. terrain), its battery level, the goal of the mission, the health of the system or the risk level that the mission managers are willing to take. To handle these challenges we rely on the risk-aware planning framework \ac{FIRM} \cite{agha2014firm}, \ac{SLAP} \cite{agha2015simultaneous} and its variant \cite{kim2019bi}.

\section{A case study of Titan}
\label{sec:MissionOperationsTitan}
\begin{figure}
\centering
\includegraphics[width=0.8\columnwidth]{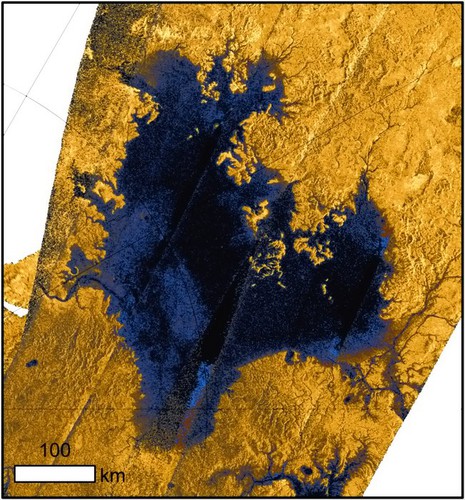}
\caption{Ligeia Mare, one of the environments on Titan accessible to the Shapeshifter. Image Source \protect\cite{esa}.}
\label{fig:LigeiaMare}
\end{figure}

\begin{figure*}[h]
\centering
\includegraphics[width=1\linewidth]{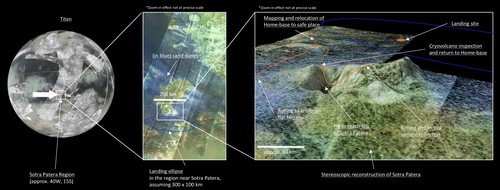}
  	\caption{Example of mission scenario near Sotra Patera. The landing area is chosen in proximity of the criovolcano Sotra Patera, and we assume a landing ellipse of 100 km by 300 km. After landing, the Shapeshifter maps the surrounding environment and relocates the Home-base to a safe place. From there, the Shapeshifter moves, by rolling or flying, to the summit of the volcano, where in-situ samples are collected. After shaping in a Rollocopter to inspect the caves near the cavity of the volcano, Shapeshifter returns to the Home-base to analyze the collected samples and plan the next mission.}
\label{fig:SotraPateraMissionArchitecture}
\end{figure*}
In this section, we present a preliminary analysis for a Titan mission scenario. 
\subsection{Science objectives at Titan and payload}

\subsubsection{Science objectives} Liquid water and organics are essential for life  but, despite being commonly found throughout the Solar System, locations where they are known to coexist are rare. Titan’s cryovolcanic regions are high priority locations to search for contact between liquid water and complex organic material because those environments would be habitable for the period of liquid water persistence. The Sotra Patera region, represented in Figure \ref{fig:SotraPateraMissionArchitecture}, is the strongest candidate cryovolcanic feature on Titan \cite{lopes2013cryovolcanism}. Shapeshifters will explore the Sotra Patera region to confirm its cryovolcanic origin and determine the extent that liquid water lavas have interacted with organic surface materials.  Shapeshifter's underwater capabilities will also allow it to explore under-liquid environments such as Ligeia Mare, represented in Figure \ref{fig:LigeiaMare}.
\subsubsection{Science payload} Each individual Cobot will carry an optical camera for the purposes of navigation and scientific imaging of Titan’s morphologies. The Cobots’ ability to image from the surface to high altitudes allows significant flexibility in image resolution and coverage. Due to size and energy limitations, the Cobots may carry only small and low-power instrumentation, such as the equipment typical considered for small spacecraft/rover hybrids (e.g. \cite{pavone2013spacecraft}), which include heat flow probes, accelerometers, magnetometers, seismometers, microphones and PH sensors.
Each or some Cobots are additionally equipped with a sample collecting unit, to collect samples of rocks or liquids to be analyzed in the Home-base. Simple technologies compatible with the Cobot's design include tongs, scoops, rakes, as extensively used during the Apollo missions, attached to the frame and actuated by the Cobot.
The Home-base will host most of the scientific payload. 
These instruments include a mass spectrometer, an X-Ray and Raman spectormeter, for analyzing the Shapeshifter samples of Titan’s complex organic materials. Using the Shapeshifter for sample collection and the Home-base for in-situ analysis is an optimal solution because the Shapeshifter can access any terrain while the in-situ instrumentation is not subject to the rolling and accelerations required for acquiring difficult samples. This combination allows for unconstrained sample acquisition and more sensitive sample analysis. 
\subsection{Exploration of Sotra Patera and other unique Titan's features} 
Our preliminary landing location is near Sotra Patera, the most likely site of cyrovolcanism on Titan, where our portable Home-base acting as an Earth relay and science laboratory, and a Shapeshifter, as a collection of 12 or more Cobots, are deployed. Once deployed, Shapeshifter will begin to morph into the optimal configuration based on the observed properties of the terrain. It will start by building high-resolution terrain maps of the region near its base. Then, it will continue shapeshifting to traverse on the Titan’s surface, gather and relay scientific data to the Home-base. Examples of the science capabilities of the platform are:
\begin{itemize}
    \item \textbf{Low/high resolution local mapping}: Shapeshifter builds maps of the region near their base;
    \item \textbf{Stratigraphy, fault survey and and surface conductivity survey}: Shapeshifter explores cliffs and faults to analyze their potential sedimentary nature and measure the conductivity of the surface;
    \item \textbf{Deep excursion}: Shapeshifter explores to its maximum range, making most efficient use of available energy by switching between the mapper and rollocopter modes;
    \item \textbf{Cave exploration}: Shapeshifter explores detected caves and cryolava tubes in the Rollocopter mode;
    \item \textbf{Mare diving for bathymetry and composition survey}: Shapeshifter morphs into a swimmer to dive under the surface of Titan’s mare, collecting samples and creating a 3D map of the surrounding environment;
    \item \textbf{Active/passive seismometry}: Basic seismography studies can be performed using the on-board accelerometers used for GN\&C, while the Home-base can host a seismometer.
\end{itemize}
After observing and analyzing a science site, Shapeshifter rebases, i.e., it morphs into a transporter and move the Home-base to a new mission site. An example of our mission is represented in Figure \ref{fig:SotraPateraMissionArchitecture}.
\subsection{Mission duration}
We envision a mission of the total duration of two Titan days, equivalent to approximately 31 Earth days. 
\subsection{Communication}
Once deployed, the Cobots will navigate the terrain by rolling, flying, and swimming all while establishing a mesh communication networks. Telemetry, in-situ measurements, and images are passed on to the home-base, which acts as a relay to get data back to Earth.

\section{Conclusions and future works}
\label{sec:ConclusionAndFutureWorks}
In this work, we have presented a novel, multi-agent and multi-modal robotic system and mission architecture for the exploration of Titan. The proposed robotic platform is capable of flying, rolling and swimming by leveraging the concept of shape-shifting, achieved by attaching and detaching the simple and affordable agents, inspired to quadcopters, of which the platform is composed. Thanks to the multi-agent and redundant nature of the system, higher risk exploration activities can be more easily negotiated. In addition, Shapeshifter's morphing capabilities allow to significantly extend the autonomy of the platform, in terms of traveled distance and types of environment explored, such as caves, cliffs, liquid basins and the cryovolcano Sotra Patera, main target of our mission on Titan. 
As part of preliminary feasibility analysis of the concept,  we have developed a simplified, two-agents prototype capable of flying, docking, rolling and un-docking, showcasing the feasibility of key design and mobility aspects of the system. From modeling and simulation results we have additionally demonstrated that the morphing capabilities of the robot indeed offer advantages in terms of increased range, as rolling on flat terrains can be up to two times more energy-efficient than flying. \newline
Future work can proceed in the following directions:
\begin{itemize}
    \item  \textbf{Multi-agent autonomy}: The robustness of a multi-agent system can be better exploited by developing distributed autonomy strategies and strategies that leverage the morphing capabilities of the robot. Examples include a distributed control system for rolling and a motion planner capable of making full use of the morphing abilities of the robot according to the traversability properties of the terrain. 
    \item \textbf{Subsystem definition}: Identify power, thermal and communication equipment compatible with the scope of the mission and the requirements of the platform. Further evaluate and define mechanical subsystems for sample collection, under-liquid docking (\textit{Torpedo mode}), under-liquid operations, and docking mechanism to transport the Home-base. Evaluate under-liquid communication strategies.
    \item \textbf{Adaptation abilities vs implementation complexity trade-offs}: Define the optimal trade-off between the morphing abilities of the platform and the increased complexity in term of development and testing.
\end{itemize}

\acknowledgments
The authors would like to thank Issa A.D. Nesnas, Kalind Carpenter, Rashied Baradaran Amini, Arash Kalantari, Jason Hofgartner, Ben Hockmann, Jonathan I. Lunine, Alessandra Babuscia, Benjamin Morrell, Jose Mendez and Kevin Liu for their precious contributions. Andrea Tagliabue thanks the \textit{Ermenegildo Zegna Founder’s Scholarship} for supporting this project. 
The research was partially carried out at the Jet Propulsion Laboratory, California Institute of Technology, under a contract with the National Aeronautics and Space Administration. The information presented about the Shapeshifter mission concept is pre-decisional and is provided for planning and discussion purposes only.
\textcopyright 2020 California Institute of Technology. Partial Government sponsorship acknowledged.

{\footnotesize
\bibliographystyle{IEEEtran}
\bibliography{bib/main.bib}
}

\thebiography
\begin{biographywithpic}
	{Andrea Tagliabue}{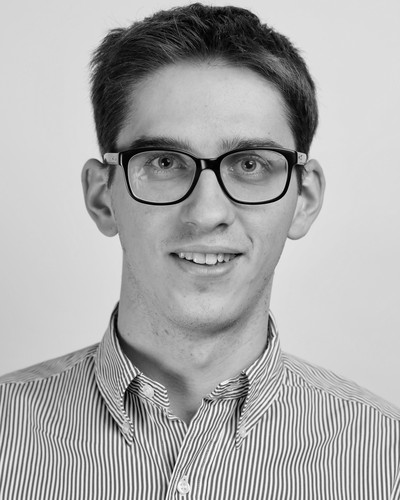} is a graduate student with the Laboratory for Information and Decision Systems at Massachusetts Institute  of  Technology. Prior to that, he was a Robotics Engineer at California Institute of Technology/NASA's Jet Propulsion Laboratory. He received a B.S. with honours in Automation Engineering from Politecnico di Milano, and a M.S. in Robotics, Systems and Control from ETH Zurich. During his master, he was visiting researcher at U.C. Berkeley and research assistant at the ETH’s Autonomous Systems Lab. His research interests include motion planning, control, and localization for Micro Aerial Vehicles. 
\end{biographywithpic}

\begin{biographywithpic}
{Stephanie Schneider}{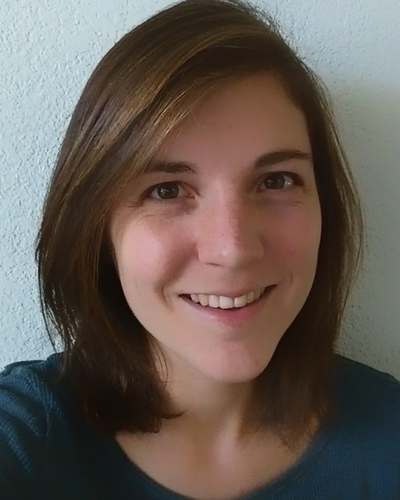}
is a graduate student with the Autonomous Systems Lab in the Department of Aeronautics and Astronautics at Stanford University. She received her BS in Mechanical Engineering from Cornell University in 2014. Prior to coming to Stanford, she worked as a software engineer and flight test engineer for Kitty Hawk Corporation (formerly Zee Aero). Her research interests include real-time spacecraft motion-planning, grasping and manipulation in space, and unconventional space robotics. 
\end{biographywithpic} 

\begin{biographywithpic}
{Dr. Marco Pavone}{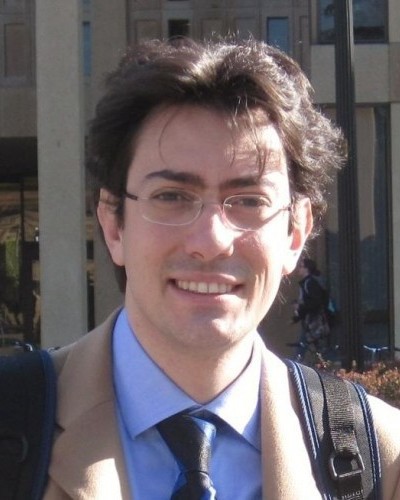}
is an Associate
Professor of Aeronautics and Astronautics at Stanford University, where he is
the Director of the Autonomous Systems
Laboratory. Before joining Stanford,
he was a Research Technologist within
the Robotics Section at the NASA Jet
Propulsion Laboratory. He received a
Ph.D. degree in Aeronautics and Astronautics from the Massachusetts Institute
of Technology in 2010. His main research interests are in
the development of methodologies for the analysis, design,
and control of autonomous systems, with an emphasis on
self-driving cars, autonomous aerospace vehicles, and future
mobility systems. He is a recipient of a number of awards,
including a Presidential Early Career Award for Scientists
and Engineers, an ONR YIP Award, an NSF CAREER Award,
and a NASA Early Career Faculty Award. He was identified
by the American Society for Engineering Education (ASEE)
as one of America’s 20 most highly promising investigators
under the age of 40. He is currently serving as an Associate
Editor for the IEEE Control Systems Magazine. 
\end{biographywithpic} 

\begin{biographywithpic}
{Dr. Ali-Akbar Agha-Mohammadi}{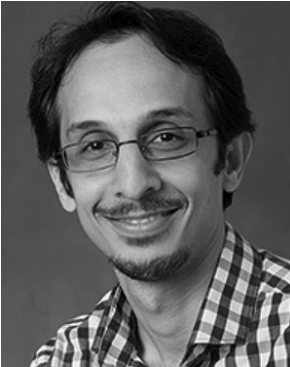} is a Robotics Research Technologist with NASA’s Jet Propulsion Laboratory (JPL), Caltech. Previously, he held a position of a Autonomy Research Engineer with Qualcomm Research and a Postdoctoral Researcher with the Laboratory for Information and Decision Systems at Massachusetts Institute of Technology. He received his Ph.D. from Texas A\&M University, College Station, TX. His research interests include robotic autonomy, mobility and perception, stochastic control systems, and filtering theory. Dr. Agha named as a NASA NIAC Fellow in 2018.
\end{biographywithpic} 

\begin{acronym}
\acro{CoM}{Center of Mass}
\acro{TRL}{Technology Readiness Level}
\acro{FIRM}{Feedback-based Information Road-Map}
\acro{SLAP}{Simultaneous Localziation and Planning}
\acro{MAV}{Micro Aerial Vehicle}
\acro{TSMP}{Traveling Sales Man Problem}
\acro{VTOL}{Vertical Take Off and Landing}
\acro{ICRA}{International Conference in Robotics and Automation}
\acro{EM}{electro-magnet}
\acro{RTG}{radioisotope termoelectric generator}
\acro{PEM}{permanent electro-magnet}
\end{acronym}

\end{document}